\documentclass{article}
\usepackage{graphicx}
\usepackage{svg}
\usepackage{subfigure}
\usepackage{ijcai21}

\usepackage{times}
\usepackage{soul}
\usepackage{url}
\usepackage[hidelinks]{hyperref}
\usepackage[utf8]{inputenc}
\usepackage[small]{caption}
\usepackage{graphicx}
\usepackage{amsmath}
\usepackage{amsthm}
\usepackage{booktabs}
\usepackage{algorithm}
\usepackage{algorithmic}
\usepackage{multirow}
\usepackage{bm}
\usepackage{amssymb}
\usepackage{pifont}
\title{SalientSleepNet: Multimodal Salient Wave Detection Network for Sleep Staging}

\author{
 Ziyu Jia$^{1,2}$\and
 Youfang Lin$^{1,2,3}$\and
 Jing Wang$^{1,2,3}$\footnote{corresponding author}\and
 Xuehui Wang$^{1}$\and
 Peiyi Xie$^{1}$\And
 Yingbin Zhang$^{1}$\\
 \affiliations
 $^1$School of Computer and Information Technology, Beijing Jiaotong University, Beijing, China\\
 $^2$Beijing Key Laboratory of Traffic Data Analysis and Mining, Beijing, China\\
 $^3$CAAC Key Laboratory of Intelligent Passenger Service of Civil Aviation, Beijing, China\\
 \emails
 \{ziyujia, yflin, wj, xuehuiwang, peiyixie, yingbinzhang\}@bjtu.edu.cn
 }

\begin{document}
	
	\maketitle
	\begin{abstract}
		Sleep staging is fundamental for sleep assessment and disease diagnosis. Although previous attempts to classify sleep stages have achieved high classification performance, several challenges remain open: 1) How to effectively extract salient waves in multimodal sleep data; 2) How to capture the multi-scale transition rules among sleep stages; 3) How to adaptively seize the key role of specific modality for sleep staging. To address these challenges, we propose SalientSleepNet, a multimodal salient wave detection network for sleep staging. Specifically, SalientSleepNet is a temporal fully convolutional network based on the $\rm U^2$-Net architecture that is originally proposed for salient object detection in computer vision. It is mainly composed of two independent $\rm U^2$-like streams to extract the salient features from multimodal data, respectively. Meanwhile, the multi-scale extraction module is designed to capture multi-scale transition rules among sleep stages. Besides, the multimodal attention module is proposed to adaptively capture valuable information from multimodal data for the specific sleep stage. Experiments on the two datasets demonstrate that SalientSleepNet outperforms the state-of-the-art baselines. It is worth noting that this model has the least amount of parameters compared with the existing deep neural network models.
	\end{abstract}
	
   	\section{Introduction}
Sleep staging is important for the assessment of sleep quality and the diagnosis of sleep disorders. To determine sleep stages, sleep experts use electrical activity recorded from sensors attached to different parts of the body. The recorded signals from these sensors are called polysomnography (PSG), consisting of electroencephalogram (EEG), electrooculogram (EOG), and other physiological signals. These recorded signals are divided into 30s sleep epochs and sleep experts classify them into five different sleep stages (W, N1, N2, N3, and REM) according to the American Academy of Sleep Medicine (AASM) sleep standard \cite{iber2007aasm}. However, this manual approach is labor-intensive and time-consuming due to the need for PSG recordings from several sensors attached to subjects over several nights. Therefore, many researchers attempt to develop automatic sleep staging methods.

\begin{figure}[t]
		\centering
		\includegraphics[width=70mm]{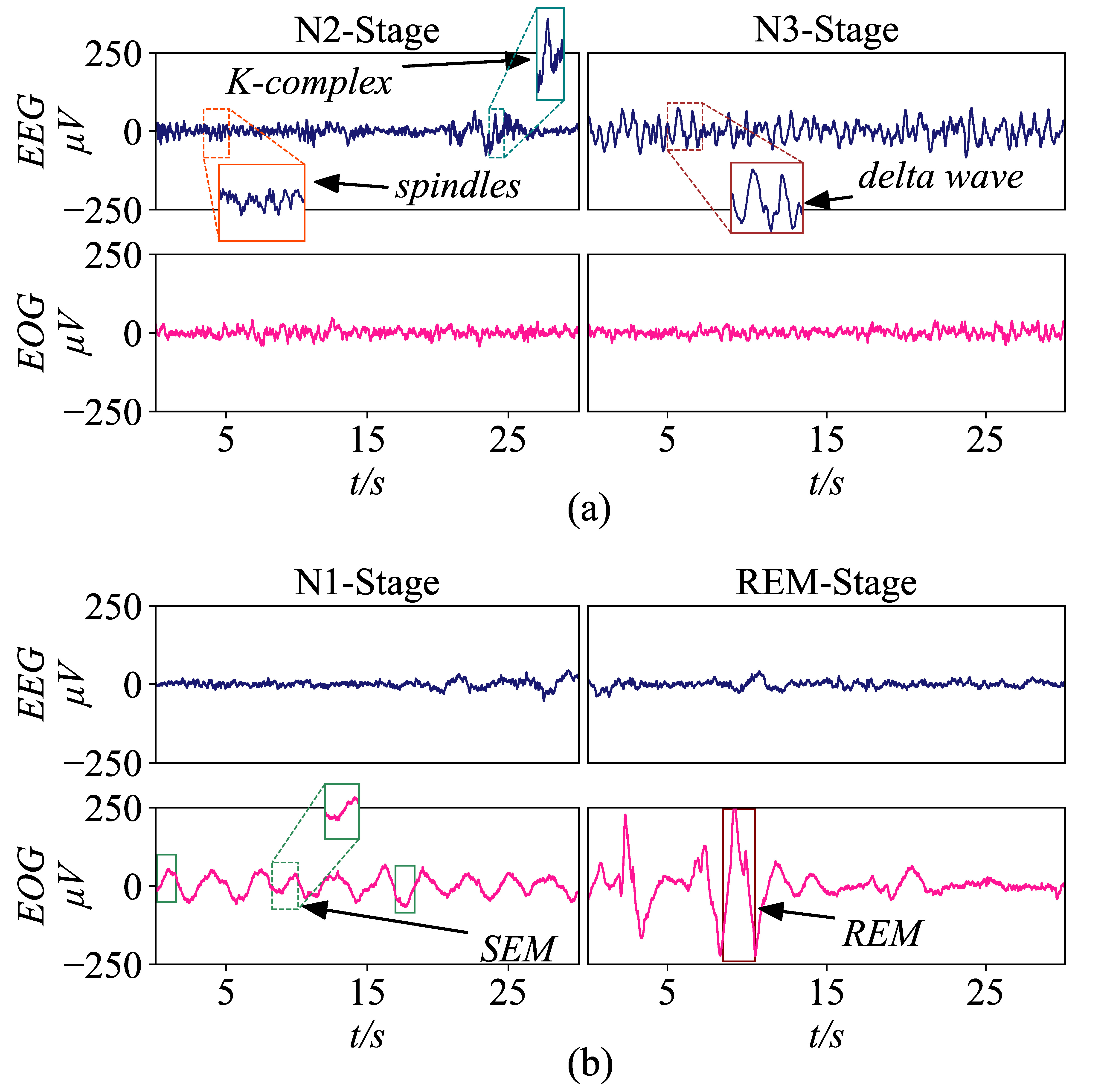}
		\caption{Salient waves in EEG and EOG signals. (a) Spindle wave and K-complex are in N2 stage. Delta wave is in N3 stage. (b) Slow eye movement (SEM) wave is in N1 stage. Rapid eye movement (REM) wave is in REM stage.}
		\label{fig:show-salient}
\end{figure}

    \begin{figure}[t]
		\centering
		\includegraphics[width=70mm]{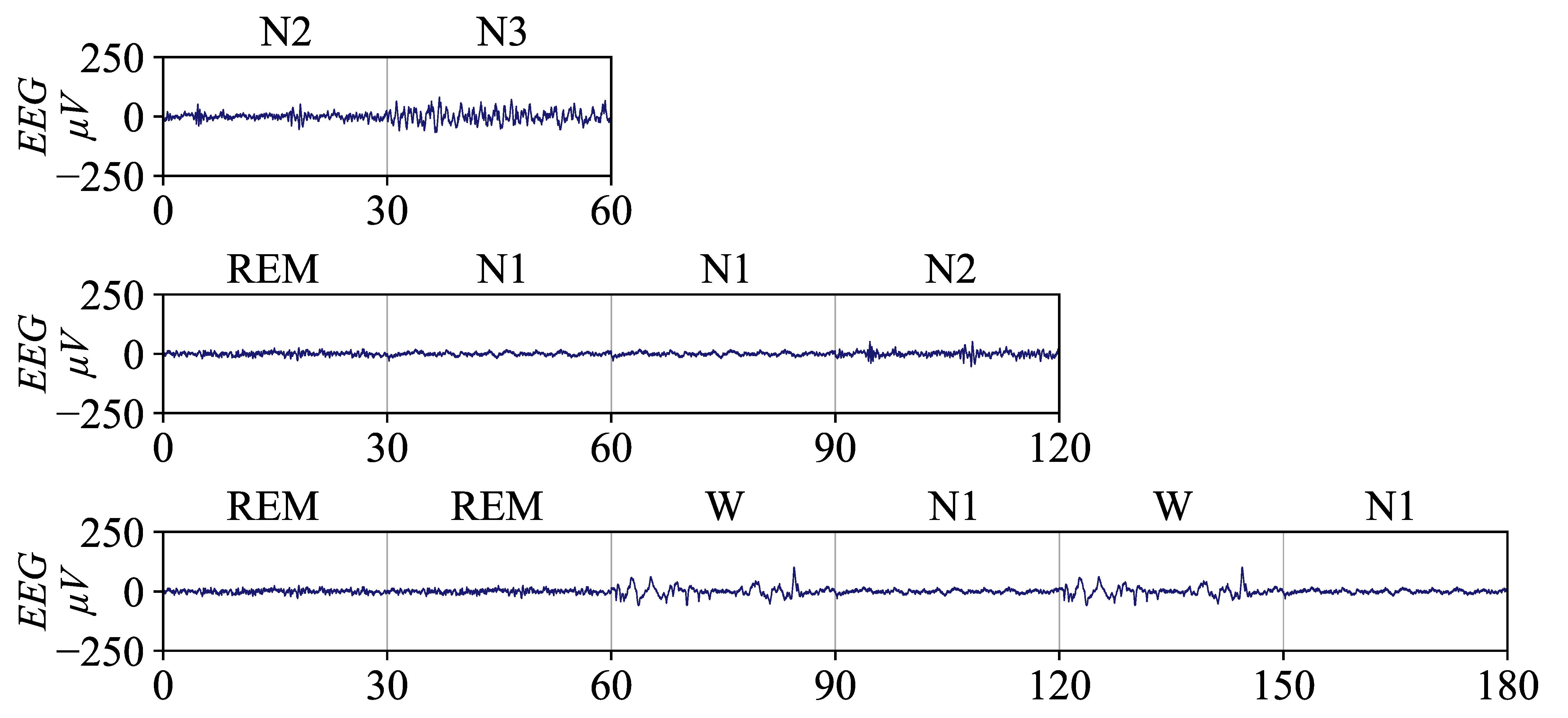}
		\caption{Different scales of sleep transition rules in AASM sleep standard.
		Short scale: $N2-N3$;
		Middle scale: $REM-N1-N1-N2$;
		Long scale: $REM-REM-W-N1-W-N1$.
		}
		\label{fig:multi-scale}
	\end{figure}



To automate sleep staging, some deep learning methods are applied to achieve state-of-the-art performance \cite{phan2019seqsleepnet,perslev2019u,supratak2020tinysleepnet}. However, there are several challenges remain open: 

1) \textit{The salient waves are directly captured from the raw signals}.
    According to the AASM sleep standard, different sleep stages usually have different salient waves in physiological signals. For example, Figure \ref{fig:show-salient} shows that the salient waves of N2 stage are spindle waves and $K$-complex waves, while the salient wave of N3 stage is delta waves. Most of the existing works use feature extraction to indirectly capture salient waves \cite{phan2019seqsleepnet,phan2020xsleepnet,chriskos2019automatic,hassan2016decision}. For example, the time-frequency features of the signals can reflect the saliency wave features to a certain extent. The raw physiological signals are converted into time-frequency images as input to deep learning models for sleep staging \cite{phan2020xsleepnet,hassan2016decision}. Although the above methods indirectly capture signal wave features, manually extracted features not only require prior knowledge but may also cause partial information loss.
    
2) \textit{Multi-scale sleep transition rules are utilized effectively.}
During sleep, the human brain undergoes a series of changes among different sleep stages. These changing patterns are summarized as transition rules in the sleep standard. Experts often determine the current sleep stage based on these rules, combined with its neighboring sleep stages. Figure \ref{fig:multi-scale} shows the sleep transition rules have multi-scale characteristics in the AASM standard.
To capture sleep transition rules, some mixed models including RNNs are usually used \cite{supratak2017deepsleepnet,supratak2020tinysleepnet,mousavi2019sleepeegnet,phan2020xsleepnet}. However, most of the existing works ignore to explicitly capture the multi-scale characteristics of sleep transition rules. In addition, these recurrent models are difficult to tune and optimize. It has been found that for many tasks RNNs models can be replaced by feed-forward systems (such as CNN variants) without sacrificing accuracy \cite{bai2018empirical}.

3)
\textit{Different modalities have different contributions to distinguish the sleep stages}.
Figure \ref{fig:show-salient} shows that the EEG waves in the REM and N1 stages are similar. However, the EOG waves in the two stages are quite different. Hence, the contribution of EOG signals to classify REM and N1 stages is greater than that of the EEG signals. On the contrary, the classification of N2 and N3 stages is mainly according to the salient waves in EEG signals. Therefore, when identifying different sleep stages, the required modalities are different.
To utilize these multimodal signals, researchers usually pay attention to the multimodal complementarity and merge the multimodal features through concatenate operations \cite{chambon2018deep,sokolovsky2019deep,phan2019seqsleepnet,phan2020xsleepnet}. This ignores that the contribution of each modality to the identification of specific sleep stages is different. 

To address the above challenges, we propose a novel deep learning model called SalientSleepNet, a sleep staging model based on multimodal sleep data (EEG signals and EOG signals). Different from many previous works, the SalientSleepNet can simultaneously capture the salient features of multimodal data and multi-scale sleep transition rules from the raw signals. 


Overall, the main contributions of the SalientSleepNet for sleep staging are summarized as follows:
\begin{itemize}
\item We develop the $\rm U^2$-structure stream, which is composed of multiple nested U-units, to detect the salient waves in physiological signals. It is inspired by the popular $\rm U^2$-Net originally proposed in computer vision.
\item We design a multi-scale extraction module to capture sleep transition rules. It is composed of multiple dilated convolutions with different receptive fields for capturing multi-scale rules.
\item We propose a multimodal attention module to adaptively capture valuable information from different multimodal data for the specific sleep stage.
\item Experimental results show that the SalientSleepNet achieves state-of-the-art performance. In addition, our model parameters are the least compared with existing deep neural networks for sleep staging.
\end{itemize}

\begin{figure*}[t]
		\centering
		\includegraphics[scale=0.36]{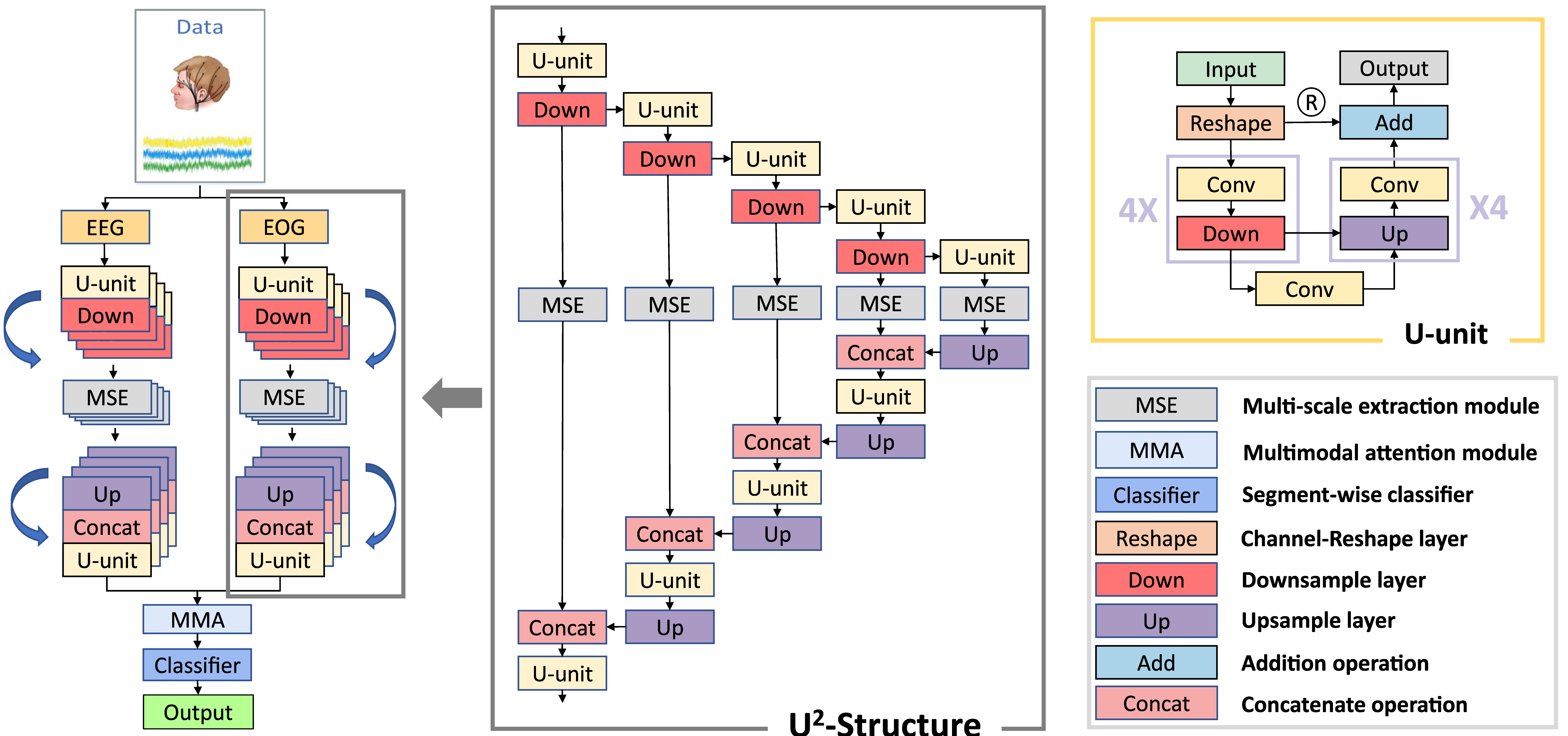}
		\caption{
		An overview architecture of SalientSleepNet. It composed of two-stream $\rm U^{2}$-structure, multi-scale feature extraction module, multimodal attention module, and the segment-wise classifier. The two-stream $\rm U^{2}$-structure extract salient wave features in EEG and EOG signal independently. Each $\rm U^{2}$-structure is composed of multiple nested U-units. U-unit has depth $l=4$. \textcircled{\tiny{R}}: the residual connection in U-unit.
		}
		\label{fig:ssn}
	\end{figure*}

\section{Related Work}

In recent years, time series analysis has attracted the attention of many researchers \cite{jia2020refined,jia2019detecting,jia2020sst}. Sleep signal as a typical physiological time series helps to diagnose sleep disorders. In earlier research, machine learning methods are utilized to classify sleep stages. However, these methods rely on hand-engineered features that require a lot of prior knowledge. Therefore, the researchers have turned to use deep learning methods for automatic sleep staging \cite{jia2020graphsleepnet,jia2021sleepprintnet,cai2020brainsleepnet}.
    
  At present, deep learning methods based on CNN or RNN have been widely used in sleep staging. For example, DeepSleepNet \cite{supratak2017deepsleepnet} utilizes CNN to extract time-invariant features, and Bi-directional Long Short-Term Memory (BiLSTM) to learn the transition rules among sleep stages. SleepEEGNet \cite{mousavi2019sleepeegnet} can extract time-invariant features from the raw signals and capture long short-term context dependencies. XSleepNet \cite{phan2020xsleepnet} presents a sequence-to-sequence sleep staging model that is capable of learning a joint representation from both raw signals and time-frequency images effectively. In addition, to lighten the existing models such as DeepSleepNet, TinySleepNet is proposed \cite{supratak2020tinysleepnet} for sleep staging. SleepUtime \cite{perslev2019u} is a fully feed-forward deep learning approach to physiological time-series segmentation. It maps sequential inputs of arbitrary length to class label sequences on a freely chosen temporal scale.
  
The existing models achieve high performance for sleep staging.  Unfortunately, these deep learning methods cannot effectively detect and fuse the salient waves in raw multimodal signals. In addition, the multi-scale transition rules among sleep stages cannot be captured explicitly.




	\section{Preliminaries}
	The proposed model processes a sequence of sleep epochs and outputs a predicted label sequence.
	Each sleep epoch is defined as $x \in \mathbb{R}^{n\times C}$, where $n$ is the number of sampling points in an epoch, and $C$ denotes the channels of sleep epoch (i.e. an EEG channel and an EOG channel in this work).
	
	The input sequence of sleep epochs is defined as $S = \{x_{1},x_{2},\ldots,x_{L}\}$, where $x_{i}$ denotes a sleep epoch ($i \in [1,2,\cdots,L]$) and $L$ is the number of sleep epochs.
	
	The sleep staging problem is defined as: learning a mapping function $F$ based on multimodal salient wave detection network which maps sleep epoch sequence $S$ into the corresponding sequence of sleep stages $\hat{Y}$, 
	where $\hat{Y} = \{\hat{y}_{1},\hat{y}_{2},\ldots, \hat{y}_{L}\}$ and $\hat{y}_{i}$ is the classification result of $x_{i}$. Following the AASM manual, each $\hat{y}_{i} \in \{0,1,2,3,4\}$ matching with the five sleep stages W, N1, N2, N3, and REM, respectively.

	\section{Multimodal Salient Wave Detection Network for Sleep Staging}
	The overall architecture of SalientSleepNet is shown in Figure \ref{fig:ssn}.
	We summarize five key ideas of SalientSleepNet: 1) Develop a two-stream $\rm U^{2}$-structure to capture the salient waves in EEG and EOG modalities. 2) Design a multi-scale extraction module by dilated convolution with different scales of receptive fields to learn the multi-scale sleep transition rules explicitly. 3) Propose a multimodal attention module to fuse the outputs from EEG and EOG streams and strengthen the features of different modalities which make greater contribution to identify certain sleep stage. 4) Improve the traditional pixel-wise (point-wise) classifier in computer vision into a segment-wise classifier for sleep signals.
	5) Employ a bottleneck layer to reduce the computational cost to make the overall model lightweight.

	\subsection{Two-Stream $\rm U^{2}$-Structure}
	Human experts classify the sleep stage mainly based on the salient waves in EEG and EOG signals such as spindle waves, K-complexes, and sawtooth waves \cite{iber2007aasm}. Existing sleep staging models extract salient wave features indirectly by converting raw signals into time-frequency images. This may cause information loss and require prior knowledge.
	
	To capture the salient waves in raw EEG and EOG signals directly, we design a two-stream $\rm U^{2}$-structure to capture the features of different signals as shown in Figure \ref{fig:ssn}. Specifically, The EEG signals and EOG signals are input into two independent $\rm U^{2}$-structures to learn the distinctive features. Each $\rm U^{2}$-structure is an encoder-decoder structure and is composed of multiple nested U-units. This is inspired by $\rm U^{2}$-Net in salient object detection of images \cite{qin2020u2}. Specifically, each U-unit has three components: a channel-reshape layer, a U-like structure, and a residual connection:
	
	
	\begin{enumerate}
		\item Given a 1D feature map $X$, the channel-reshape layer transforms it into an intermediate feature map to control the channel number of the U-unit:
		\begin{align}
		X_{m}=Reshape(X),
		\end{align}
		where $Reshape$ denotes the channel-reshape operation and $X_{m}$ denotes the intermediate feature map.
		\item The U-like structure encodes and decodes the intermediate feature map and obtains $X_{m}'$:
		\begin{align}
		X_{m}'=U_{l}(X_{m}),
		\end{align}
		where $U_{l}$ denotes the U-like structure with depth of $l$ ($l=4$ in our model).
		\item The residual connection  \cite{he2016deep} fuses feature map $X_{m}$ and $X_{m}'$ with addition operation to reduce the degradation problem in the deep network:
		\begin{align}
		X_{m}'' = X_{m} + X_{m}',
		\end{align}
		where $X_{m}''$ is the output of the U-unit.
	\end{enumerate}
	
	Multiple U-units make up the $\rm U^{2}$-structure. Specifically, there are 5 U-units in the encoder and 4 U-units in the decoder for each $\rm U^{2}$-structure.
	
	
	\subsection{Multi-scale Extraction Module}
	Sleep transition rules is important for sleep staging.
	The transition rules have the characteristics of multiple scales: short scale, middle scale, and long scale. 
    Previous works use the RNN to learn the sleep transition rules. However, the RNN models are difficult to tune and optimize \cite{bai2018empirical}. 
	
	
	\begin{figure}[ht]
		\centering
		\includegraphics[width=65mm]{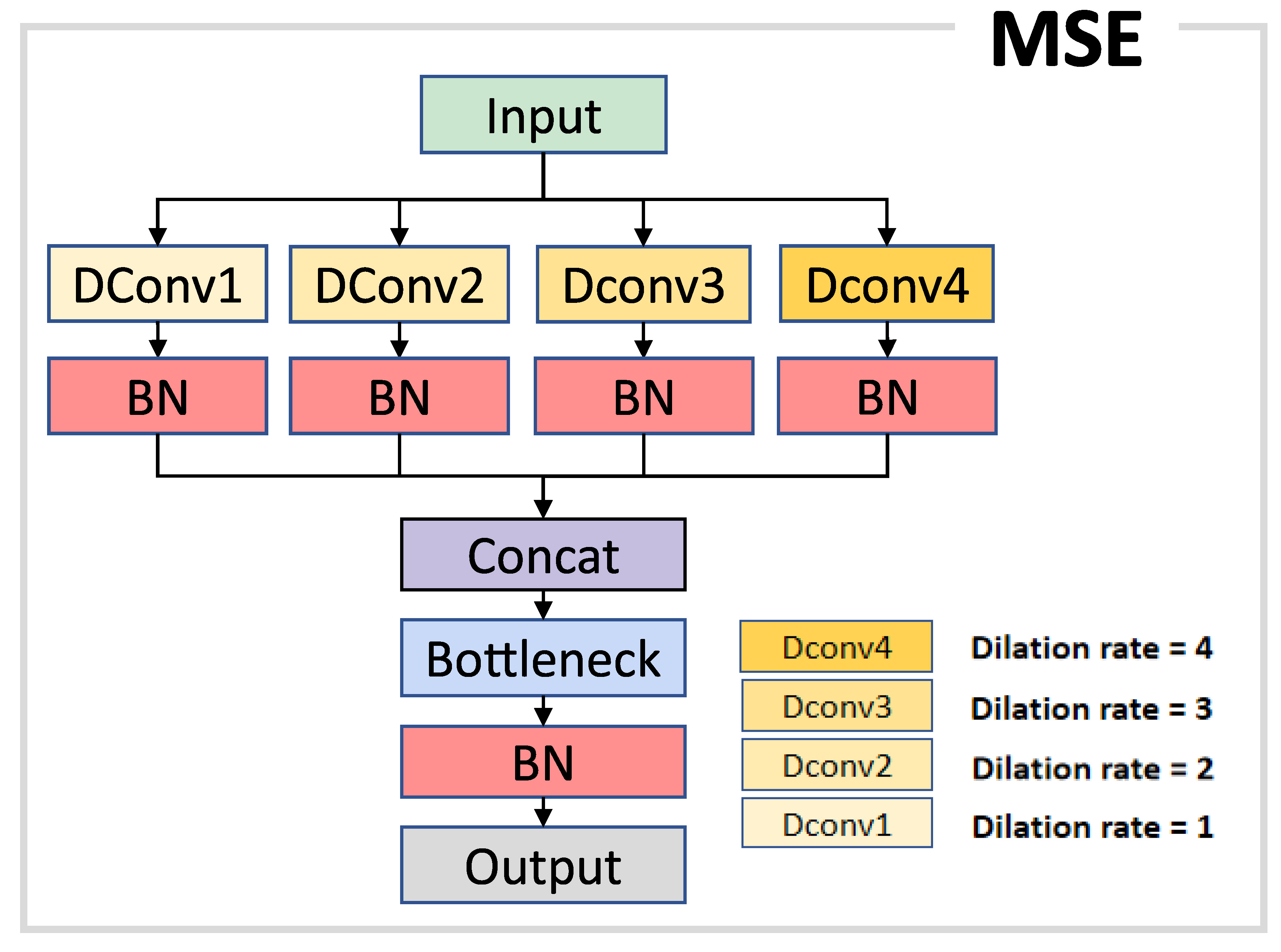}
		\caption{The structure of \textbf{M}ulti-\textbf{S}cale \textbf{E}xtraction module (MSE) which captures multi-scale transition rules of sleep epoch sequence.
		Dconv$r$ is the $r$-th dilation convolution with dilation rate $r$.
		Bottleneck: Bottleneck layer which reduce the channel of feature map.
		Concat: Concatenate operation;
		BN: Batch normalization.
		}
		\label{fig:mse}
	\end{figure}

	To solve the above problems, we design a \textbf{M}ulti-\textbf{S}cale \textbf{E}xtraction module (MSE) to capture multi-scale sleep transition rules explicitly, as shown in Figure \ref{fig:mse}. The MSE has dilated convolutions with different dilation rates to capture features in different scales of receptive fields \cite{yu2015multi}.
	Specifically, we employ 4 dilated convolutions with dilation rates from 1 to 4 to transform the same input feature map.
	Then, the feature maps learned from different scales of receptive fields are concatenated to obtain a multi-scale feature map, which is defined as:
	\begin{align}
	X_{d}^{r} = DConv_{r}(X_{m}''), r \in [1,2,3,4],
	\end{align}
	\begin{align}
	X_{ms} = Concat(X_{d}^{1},X_{d}^{2},X_{d}^{3},X_{d}^{4}),
	\end{align}
	where $X_{m}''$ is the input feature map,
	$DConv_{r}$ is the dilated convolution with dilation rate $r$,
	$X_{d}^{r}$ is the output of dilated convolution $DConv_{r}$, and $X_{ms}$ is the multi-scale feature map.
	
	Besides, to reduce the parameters of our model, we apply a bottleneck layer \cite{he2016deep} between the encoder and decoder (i.e. implemented in MSE). It reduces the channels of the concatenated feature map which makes the model lightweight and composed of two convolution operations.
	The bottleneck layer is defined as:
	\begin{align}
	X_{b} = Bottleneck(X_{ms}),
	\end{align}
	where $Bottleneck$ is the operation of bottleneck layers and $X_{b}$ is the final multi-scale feature map with channel reduction operation obtained from the MSE.
	$X_{b}$ has $C_{out}$ channels and $C_{out} = C_{in} / rate$, where $C_{in}$ is the channel number of $X_{ms}$ and $rate$ is the down-sampling rate of bottleneck layer.
	
	\subsection{Multimodal Attention Module}
	Different modalities have distinctive features which contribute to classifying specific sleep stage.
	However, existing models ignore that different modalities have different contributions to identify certain sleep stage.
	
\begin{figure}[ht]
		\centering
		\includegraphics[height=75mm,width=50mm]{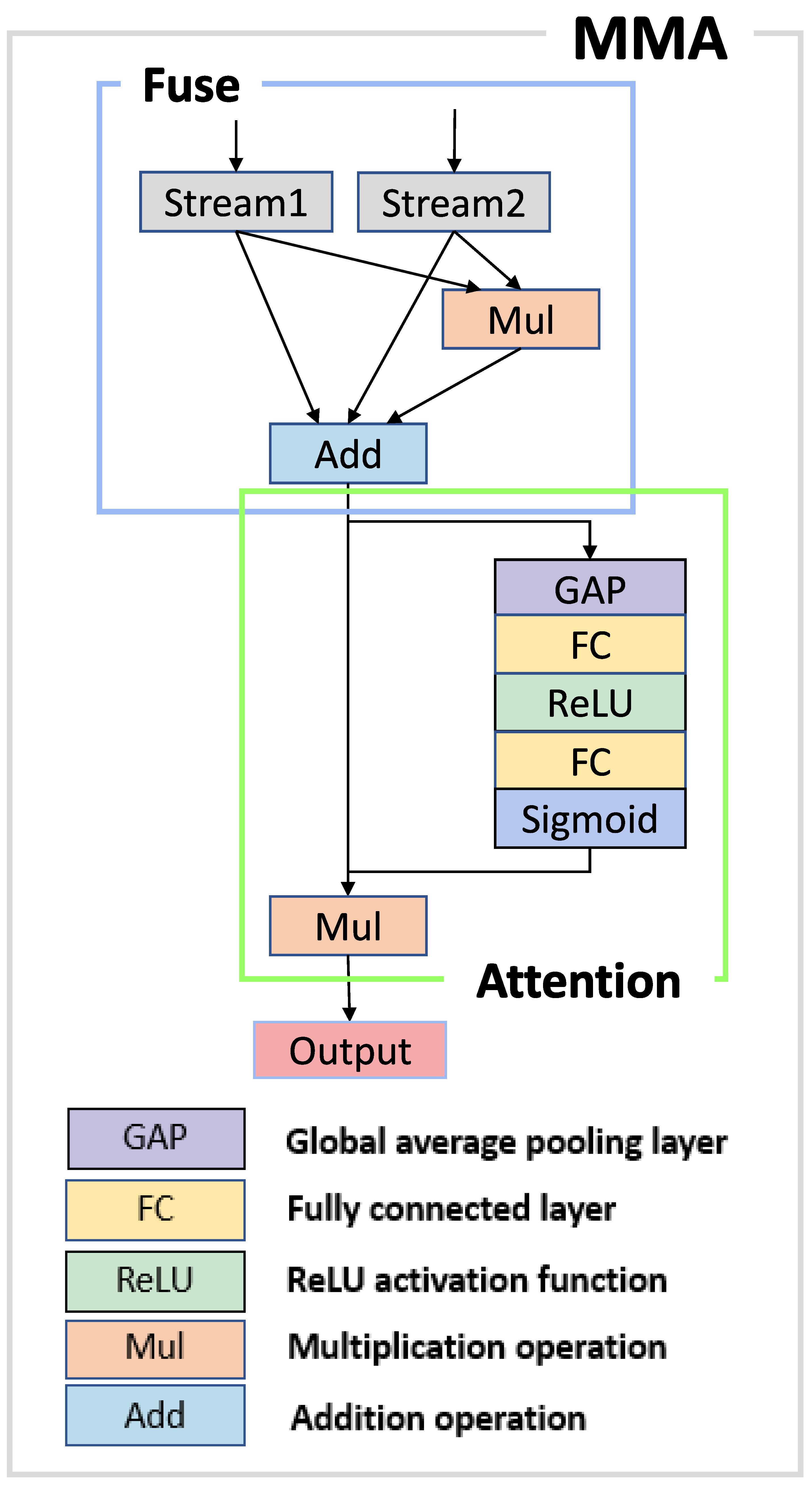}
		\caption{The structure of \textbf{M}ulti\textbf{M}odal \textbf{A}ttention module (MMA) which adaptively capture the important features of different modalities for certain sleep stage.
		}
		\label{fig:mma}
	\end{figure}

	To strengthen the features of different modalities which have greater contribution to classify specific sleep stage in an adaptive approach, we propose a \textbf{M}ulti\textbf{M}odal \textbf{A}ttention module (MMA).
	As shown in Figure \ref{fig:mma}, The MMA has two main components: The modality fusion component for fusing the feature maps from two streams and the channel-wise attention component \cite{hu2018squeeze} for strengthening the most important features in an implicit way for certain sleep stage.

	\begin{enumerate}
		\item Inspired by the fusion method of \cite{fu2020jl}, the modality fusion component is defined as:
		\begin{align}
		X_{fuse} = X_{EEG} + X_{EOG} + (X_{EEG}\odot X_{EOG}),
		\end{align}
		where $X_{fuse}$ is the fused feature map and $\odot$ is the element-wise multiplication. $X_{EEG}$ and $X_{EOG}$ are the feature maps learned from EEG and EOG streams, respectively. 
		\item The $X_{fuse}$ is input into the channel-wise attention component to strengthen the important channels of the fused feature map:
		\begin{align}
		X_{att} = \sigma(FC_{2}(\delta(FC_{1}(GAP(X_{fuse}))))),
		\end{align}
		\begin{align}
		X_{att}' = X_{fuse}\odot X_{att},
		\end{align}
		where $GAP$ denotes the global average pooling operation, $FC_{i}$ denotes $i$-th fully connected layers, $\delta$ is the ReLU activation function, $\sigma$ is the sigmoid activation function, $X_{att}$ is the intermediate feature map, and $X_{att}'$ is the output of this component.
	\end{enumerate}
	
	\subsection{Segment-wise Classification}
	Existing salient object detection models in computer vision are pixel-wise (point-wise) classification \cite{qin2020u2}. These models can not be applied directly to physiological signal segment-wise classification.
	
	Hence, we design a segment-wise classifier which maps the pixel-wise feature map to a segment-wise predict label sequence.
	Specifically, an average pooling layer is applied to reduce the size of 1D feature map from $X_{att}' \in \mathbb{R}^{L'}$ to $X_{pool} \in \mathbb{R}^{L}$, where $L'= L*n$, $L$ is the number of sleep epochs, $n$ represents the number of sampling points in an epoch.
	Then, a convolution layer with Softmax activation is used to reduce the channel dimension of $X_{pool}$ and map vectors to predict label sequence $\hat{Y}.$ Figure A.\ref{fig:swc} in Appendix shows the structure of the segment-wise classifier.

	\section{Experiments}
	We evaluate the performance of SalientSleepNet on Sleep-EDF-39 and Sleep-EDF-153 datasets \cite{kemp2000analysis,goldberger2000physiobank}.
	
	\subsection{Datasets and Preprocessing}
	\textit{Sleep-EDF-39} consists of 42308 sleep epochs from 20 healthy subjects (10 males and 10 females) aged 25-34. Each subject contains two day-night PSG recordings except subject 13 whose one recording is lost due to device failure. \textit{Sleep-EDF-153} consists of 195479 sleep epochs from 78 healthy subjects aged 25-101. Similar to Sleep-EDF-39, each subject contains two day-night PSG recordings except subjects 13, 36, and 52 whose one recording is lost due to device failure.
    
  For a fair comparison with other methods, we use the same preprocessing and data for all models. Specifically, each 30s epoch of the recordings is labeled into one of six categories \{W, N1, N2, N3, N4, REM\}. We merge N3 and N4 stages into a single N3 stage according to the latest AASM sleep standard \cite{iber2007aasm} and remove before and after in-bed epochs for only in-bed recordings are used as recommended in \cite{imtiaz2014recommendations}. We adopt Fpz-Cz EEG and ROC-LOC EOG channels in this study. All signals have a sampling rate of 100 Hz.
    

		\begin{table*}
		\resizebox{\textwidth}{!}{
        \begin{tabular}{crcccccccccccccc}
        \toprule
        \toprule

        & & \multicolumn{7}{c}{Sleep-EDF-39 dataset} & \multicolumn{7}{c}{Sleep-EDF-153 dataset} \\ \cmidrule(r){3-9} \cmidrule{10-16} 
 
        Method& Parameters& \multicolumn{2}{c}{Overall results} & \multicolumn{5}{c}{F1-score for each class} & \multicolumn{2}{c}{Overall results} & \multicolumn{5}{c}{F1-score for each class}    \\ \cmidrule(r){3-9} \cmidrule{10-16}

        & & F1-score & Accuracy & Wake & N1 & N2 & N3 & REM &
        F1-score & Accuracy & Wake & N1 & N2 & N3 & REM \\
        \cmidrule(r){1-9} \cmidrule{10-16}

        SVM & \textless0.1M & 63.7 & 76.1 & 71.6 & 13.6 & 85.1 & 76.5 & 71.8 
        & 57.8 & 71.2 & 80.3 & 13.5 & 79.5 & 57.1 & 58.7 \\

        RF & \textless0.1M & 67.6 & 78.1 & 74.9 & 22.5 & 86.3 & 80.8 & 73.3 
        & 62.4 & 72.7 & 81.6 & 23.2 & 80.6 & 65.8 & 60.8 \\

        DeepSleepNet & 21M & 76.9 & 82.0 & 85.0 & 47.0 & 86.0 & 85.0 & 82.0 
        & 75.3 & 78.5 & 91.0 & 47.0 & 81.0 & 69.0 & 79.0 \\

        SeqSleepNet & -- & 79.7 & 86.0 & 91.9 & 47.8 & 87.2 & 85.7 & 86.2 
        & 78.2 & 83.8 & 92.8 & 48.9 & 85.4 & 78.6 & 85.1 \\

        SleepEEGNet & 2.1M & 79.7 & 84.3 & 89.2 & 52.2 & 86.8 & 85.1 & 85.0 
        & 77.0 & 82.8 & 90.3 & 44.6 & 85.7 & \textbf{81.6} & 82.9 \\

        SleepUtime & 1.1M & 79.0 & -- & 87.0 & 52.0 & 86.0 & 85.0 & 82.0 
        & 76.0 & -- & 92.0 & 51.0 & 84.0 & 75.0 & 80.0 \\

        TinySleepNet & 1.3M & 80.5 & 85.4 & 90.1 & 51.4 & 88.5 & \textbf{88.3} & 84.3 
        & 78.1 & 83.1 & 92.8 & 51.0 & 85.3 & 81.1 & 80.3 \\

        \cmidrule(r){1-9} \cmidrule{10-16}

        \textbf{SalientSleepNet} & \textbf{0.9M} & \textbf{83.0} & \textbf{87.5} & \textbf{92.3} & \textbf{56.2} & \textbf{89.9} & 87.2 & \textbf{89.2}
        & \textbf{79.5} & \textbf{84.1} & \textbf{93.3} & \textbf{54.2} & \textbf{85.8} & 78.3 & \textbf{85.8} \\

        \bottomrule
        \bottomrule
        \end{tabular}
        }

		\caption{Performance comparison of the state-of-the-art approaches on Sleep-EDF-39 and Sleep-EDF-153 datasets. "--" indicates the corresponding value not provided in the baseline models.
		}
		\label{tab:result}
		
	\end{table*}

\subsection{Baseline Methods}
We compare the SalientSleepNet with the following baselines:

\begin{itemize}
\item 
SVM \cite{hearst1998support}: 
Support Vector Machines uses a Gaussian kernel function for sleep staging.
\item 
RF \cite{breiman2001random}: 
Random Forests are an ensemble learning method for sleep staging.
\item 
DeepSleepNet \cite{supratak2017deepsleepnet}: 
DeepSleepNet utilizes CNN to extract time-invariant features and BiLSTM to learn the transition rules among sleep stages. 
\item 
SeqSleepNet \cite{phan2019seqsleepnet}:
SeqSleepNet is composed of parallel filterbank layers for preprocessing the time-frequency images and bidirectional RNN to encode sleep sequential information. 
\item 
SleepEEGNet \cite{mousavi2019sleepeegnet}:
SleepEEGNet extract time-invariant features from the sleep signals and capture long short-term context dependencies 
\item 
SleepUtime \cite{perslev2019u}:
SleepUtime maps sequential inputs of arbitrary length to sequences of class labels on a freely chosen temporal scale.
\item 
TinySleepNet \cite{supratak2020tinysleepnet}:
TinySleepNet is a mixed model of CNN and unidirectional RNN.
\end{itemize}

	\subsection{Experiment Settings}
	We implement the SalientSleepNet based on the TensorFlow framework. Our model is trained by Adam optimizer with learning rate $\eta=10^{-3}$. The training epoch is 60 and batch size is 8. The input sleep epoch sequence length $L$ is 20. Besides, the down-sampling rate of the bottleneck layer $rate$ is 4. We apply 20-fold cross-validation to evaluate the cross-subject performance of SalientSleepNet. Other hyperparameters are given in Table A.\ref{tab:hyp} in Appendix and our code is publicly available at \href{https://github.com/ziyujia/SalientSleepNet}{https://github.com/ziyujia/SalientSleepNet}.

	
	\subsection{Experiment Results}
	We compare SalientSleepNet with other seven baseline methods on two publicly available datasets as shown in Table \ref{tab:result}. The results show that SalientSleepNet achieves the best overall performance compared with other baseline methods. 
	
	The traditional machine learning models (SVM and RF) can't capture various features.
	Some mixed deep learning models such as DeepSleepNet, SeqSleepNet, and TinySleepNet combine CNN to learn the features in sleep epochs and RNN to capture transition rules between sleep epochs. Therefore, These models achieve better performance than traditional machine learning models.
	
    Although the mixed models can achieve high accuracy, these models are difficult to tune and optimize. In addition, existing models do not make full use of salient wave features in different modalities and some models require time-frequency images as inputs which may cause partial information loss.  Different from previous works, our model can simultaneously capture the salient features of multimodal data and multi-scale sleep transition rules from the raw signals. In addition, our model strengthens the contribution of different modalities features to classify different sleep stages. Therefore, SalientSleepNet achieves better overall performance than other baseline methods. It is worth noting that the N1 stage usually has the smallest sample number and is easy to confuse with REM stage. However, our model improves the F1-score for N1 stage to a certain extent. Besides, our model has the minimum parameters compared with other deep learning models. 
    
\begin{figure}[ht]
		\centering
		\includegraphics[width=\columnwidth]{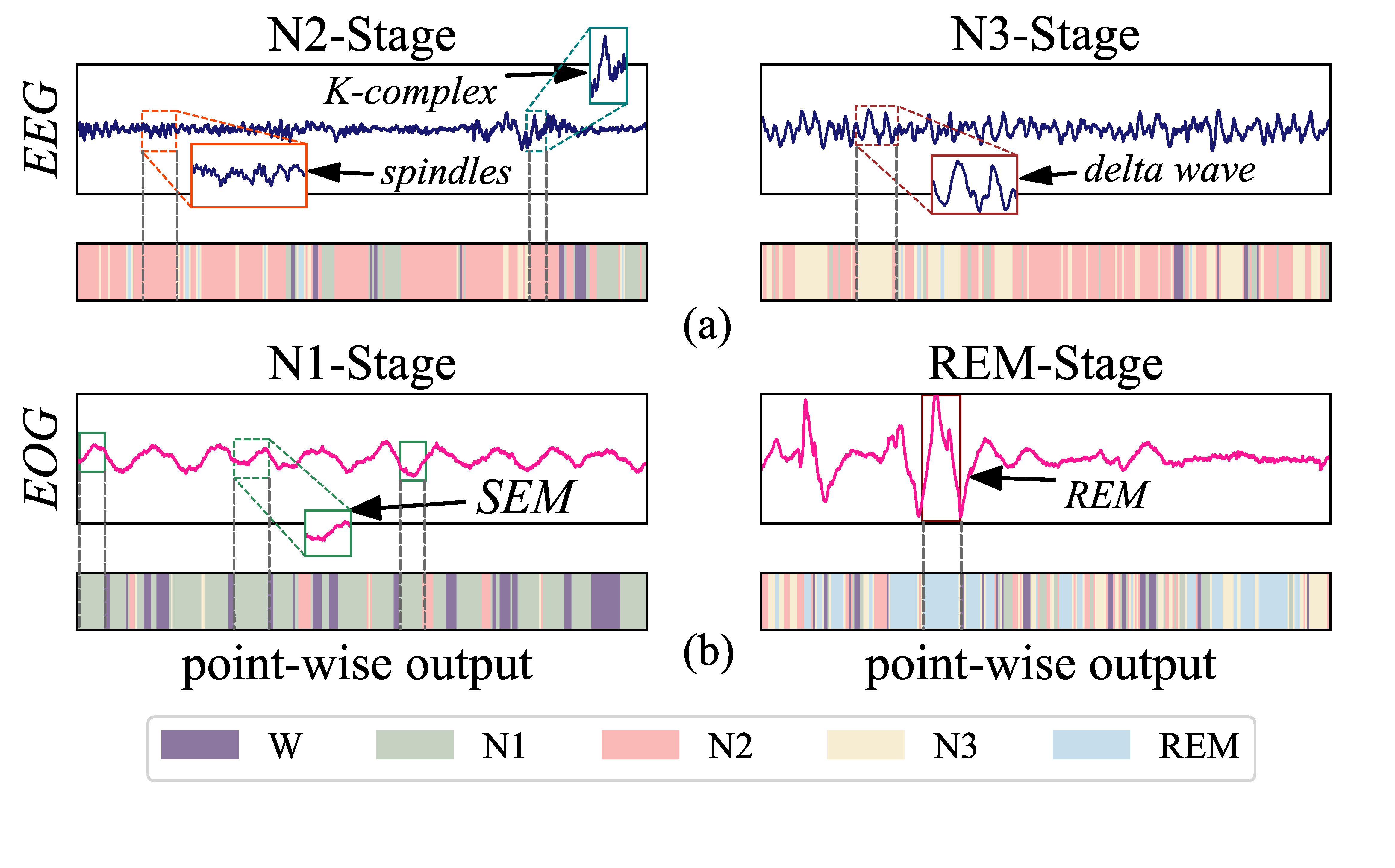}
		\caption{Salient waves detected by SalientSleepNet. (a) The detected salient waves in EEG signals for N2 and N3 stages. (b) The detected salient waves in EOG signals for N1 and REM stages.}
		\label{fig:v-salient}
    \end{figure}

	
	We visualize the point-wise outputs of $\rm U^{2}$-structure as shown in
    Figure \ref{fig:v-salient}. It illustrates that our model can detect salient waves in multimodal signals to a certain extent.
    For example, in EEG signals, the sleep spindle wave and K-complex are in N2 stage. The delta wave appears in N3 stage.
	Besides, in EOG signals, the slow eye movement wave and rapid eye movement wave exist in N1 stage and REM stage, respectively. Overall, our model can detect these salient waves, which makes this model interpretable to a certain extent.
	
	\subsection{Ablation Experiment}
	To evaluate the effectiveness of each module in our model, we design several variant models:
	\begin{itemize}
		\item U structure (basic): This model is a two-stream U structure without nested U-units, MSE, and MMA. The feature maps obtained from two streams are concatenated for sleep staging.
		\item $\rm U^{2}$ structure: The nested U-units are applied to the basic model.
		\item $\rm U^{2}$ structure+MSE: This model adds the multi-scale extraction modules based on $\rm U^{2}$ structure.
		\item $\rm U^{2}$ structure+MSE+MMA (SalientSleepNet): This model adds the multimodal attention module instead of a simple concatenate operation based on $\rm U^{2}$+MSE model.

	\end{itemize}
	\begin{figure}[ht]
		\centering
		\includegraphics[width=\columnwidth]{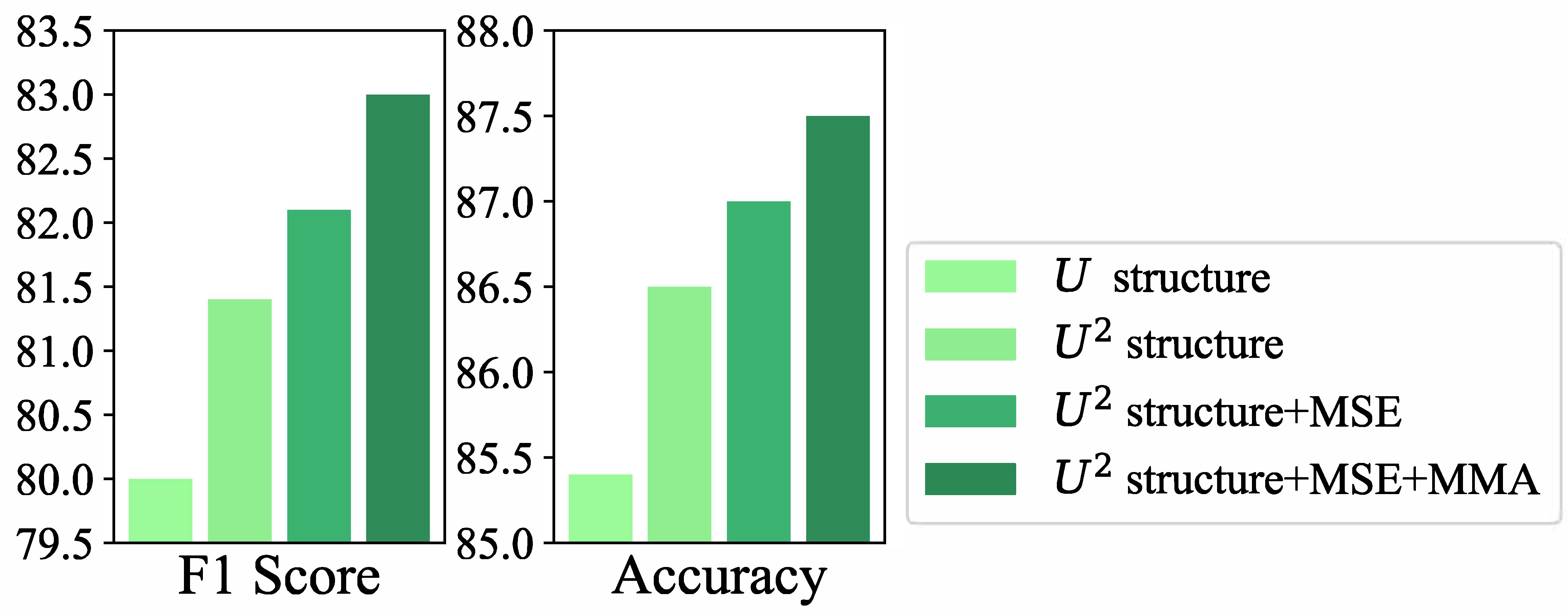}
		\caption{The results of ablation experiment.}
		\label{fig:ablation}
	\end{figure}
    
	Figure \ref{fig:ablation} presents the $\rm U^{2}$-structure obviously improves the performance of basic model, because it captures more diverse salient waves. Then, the structure with MSE learns the multi-scales features which improve the performance. In addition, the MMA is added to adaptive capture the important features of different modalities for certain sleep stage.
	
	Besides, to evaluate the effectiveness of the bottleneck layer in reducing model parameters, we removed this layer in MSE. Compared with the model without bottleneck layer which has 1,128,865 parameters, SalintSleepNet only has 948,795 parameters. The bottleneck layer reduces the number of parameters in the model by $16\%$. It allows our model to keep the smallest size while achieving the best performance than other baseline methods at the same time.
	
	
    

	\section{Conclusion}
	In this paper, we propose a multimodal salient wave detection network for sleep staging. Our work is the first attempt to borrow the $\rm U^{2}$-Net from the visual saliency detection domain to sleep staging. SalientSleepNet can not only effectively detect and fuse the salient waves in multimodal data, but also extract multi-scale transition rules among sleep stages. Experiment results show that SalientSleepNet achieves state-of-the-art performance. The parameters of our model are the least among the existing deep learning models. In addition, the proposed model is a general-framework for multimodal physiological time series.
	
    \section*{Acknowledgments}
This work was supported by the Fundamental Research Funds for the Central Universities (Grant No.2020YJS025) and the China Scholarship Council (Grant No.202007090056).

	\bibliographystyle{named}
	{\small
	\bibliography{ijcai21}}
	
	\newpage
	\appendix 
	\section*{Appendix}
	\captionsetup[figure]{name={Figure A.},font={large}}
    \captionsetup[table]{name={Table A.},font={large}}
	\setcounter{table}{0}
	\setcounter{figure}{0}
	
	\begin{table}[!h]
	\caption{\mbox{The description of all sleeps stages and their salient waves.}}
	\centering
		\begin{tabular}{lp{40pt}p{300pt}}
		
			\toprule
			\textbf{Name} & \textbf{Encoding} & \textbf{Description} \\
			\midrule
			Non-REM 1 & N1 & Short, light sleep stage comprising about 5\%-10\% of a night’s sleep. Dominated by theta waves (4-7 Hz EEG signals). Slow eye movements in W→ N1 transition. Some EMG activity, but lower than wake. \\
			Non-REM 2 & N2 & Comprises 40\%-50\% of a normal night’s sleep. EEG displays theta-waves like N1, but intercepted by so-called K-complexes and/or sleep spindles (short bursts of 13-16Hz EEG signal). \\ 
			Non-REM 3 & N3 & Comprises about 20\%-25\% of a typical night’s sleep. High amplitude, slow 0.3-3 Hz EEG signals. Low EMG activity. \\
			REM & REM & Rapid-eye-movements may occur. Displays both theta waves and alpha (like wake), but typically 1-2 Hz slower. EMG significantly reduced. Dreaming may occur this stage, which comprises 20\%-25\% of the night’s sleep.\\
			Wake & W & Spans wakefulness to drowsiness. Consists of at least 50\% alpha waves (8-13 Hz EEG signals). Rapid and reading eye movements. Eye blinks may occur. \\
			\bottomrule
		\end{tabular}	
	
	\label{tab:sleep-satge}	
	\end{table}
	
	\begin{table}[!h]
    \centering
    \caption{\mbox{Sample number and sample proportion in Sleep-EDF-39 and Sleep-EDF-153 datasets.}}
		\resizebox{100mm}{!}{
			\begin{tabular}{r|rr|rr}
				\toprule
				\multirow{2}{*}{\textbf{Label}}
				& \multicolumn{2}{c|}{Sleep-EDF-39} 	

& \multicolumn{2}{c}{Sleep-EDF-153} \\\cmidrule{2-5}
				 & \textbf{Number} & \textbf{Proportion} & \textbf{Number} & \textbf{proportion} \\
				\midrule
				
				W & 8285 & 19.58\% & 65951 & 33.74\% \\
				N1 & 2804 & 6.63\% & 21522 & 11.01\% \\
				N2 & 17799 & 42.07\% & 69132 & 35.37\% \\
				N3 & 5703 & 13.48\% & 13039 & 6.67\% \\
				REM & 7717 & 18.24\% & 25835 & 13.22\% \\
				\midrule
				\textbf{total} & \textbf{42308}& \textbf{100.00\%}& \textbf{195479}& \textbf{100.00\%}\\
				\bottomrule
			\end{tabular}
			}
		
		\label{tab:labels} 	
		
	\end{table}
	
	\begin{figure*}[!htb]
		\centering
		\includegraphics[width=\textwidth]{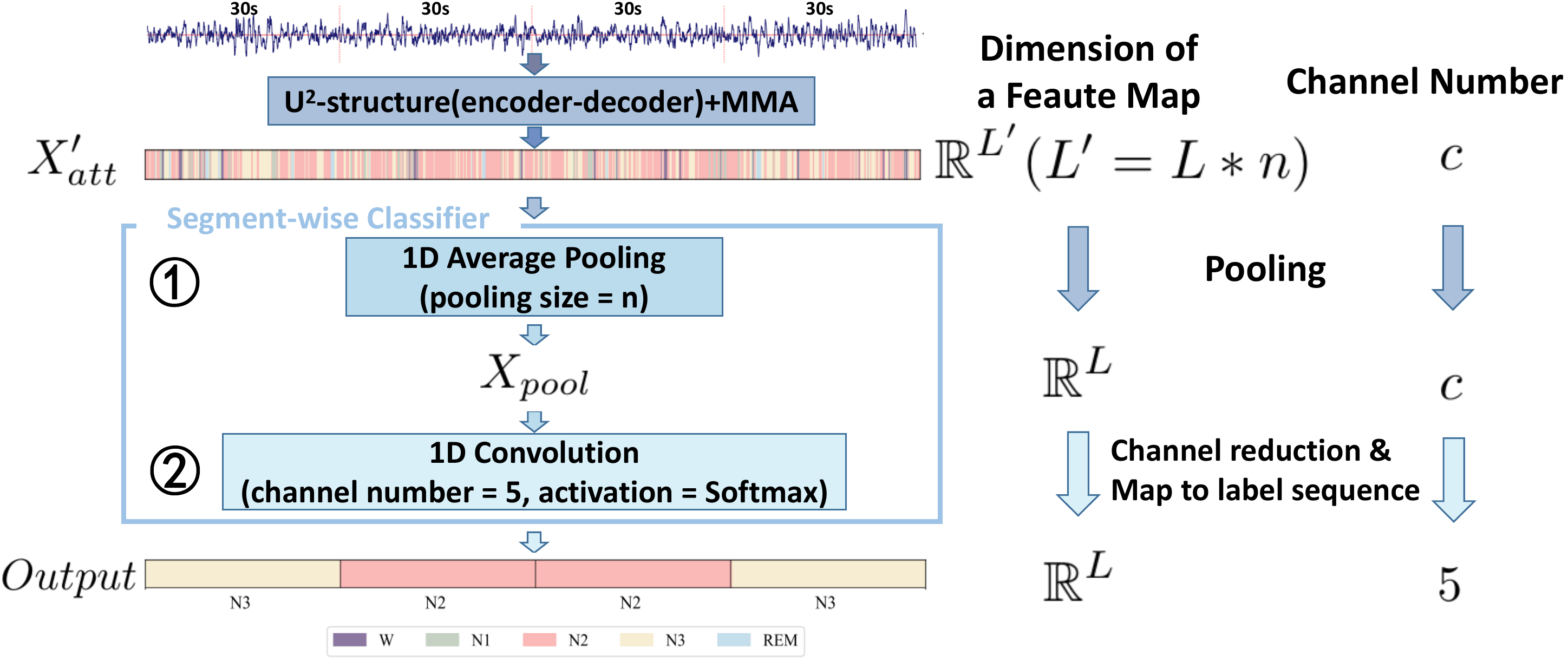}
		\caption{The operation of the segment-wise classifier(SWC) which serves as a trainable link between the intermediate representation defined by the encoder-decoder network and the label space. SWC drives the encoder-decoder network to detect salient waves. Specifically, \ding{172} $X'_{att}$ is input to an average pooling layer with pooling size $n$, which reduces the length of $X'_{att}$ to $L$ and forms $X_{pool}$.
\ding{173} We follow the output format of the fully convolutional architecture. A convolution layer with Softmax activation are used to reduce the channel dimension and map vectors to $\hat{Y}$.}
		\label{fig:swc}
	\end{figure*}
	
	\begin{figure*}[!h]
		\centering
		\includegraphics[width=\textwidth]{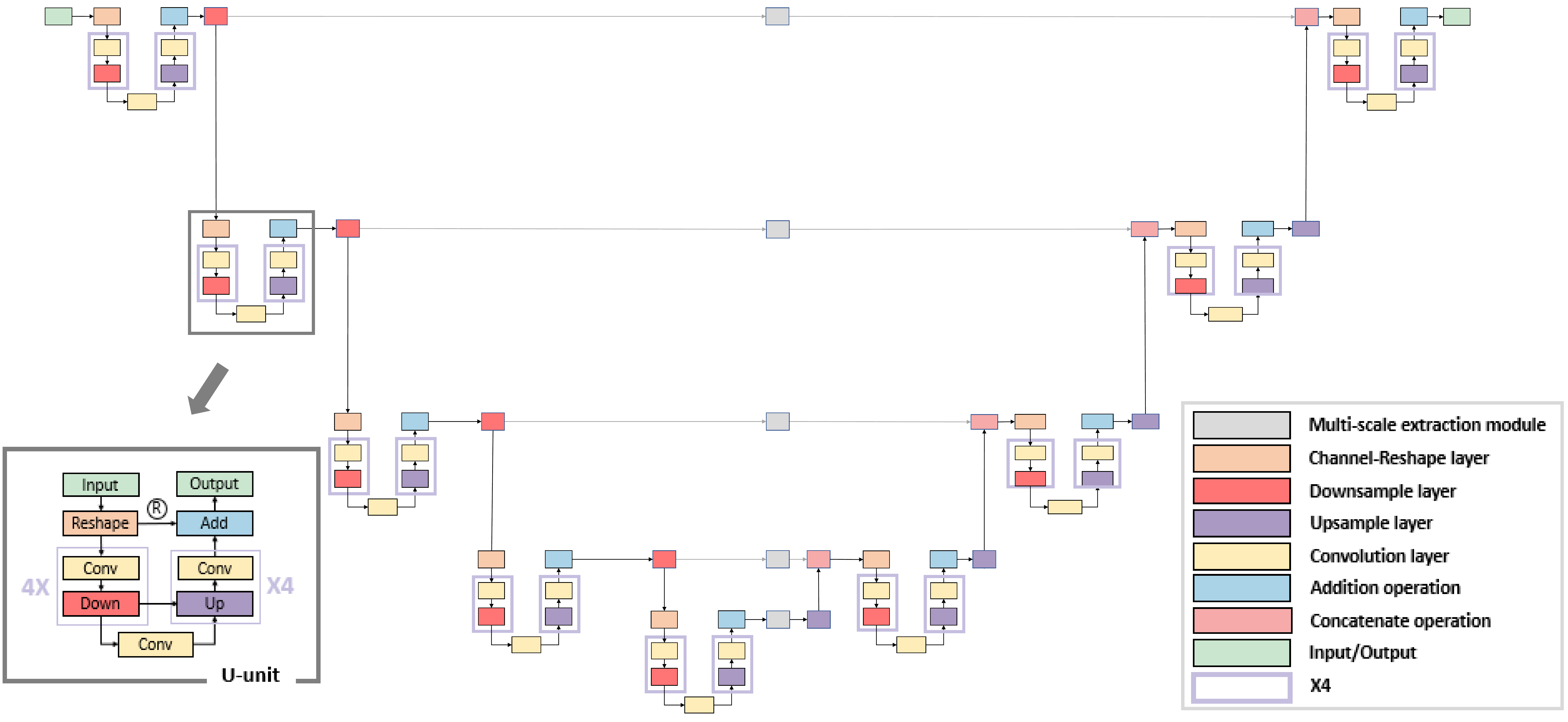}
		\caption{The overview of $\rm U^2$-structure.}
		\label{fig:u2}
	\end{figure*}
	
	\clearpage
	
	\begin{figure}[!t]
	\centering    
	
	{
		\begin{minipage}{\columnwidth}
   			\includegraphics[width=\columnwidth]{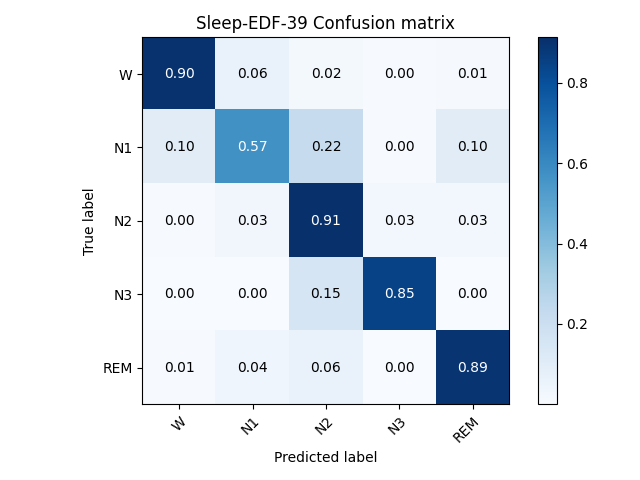}
		\end{minipage}
	}
	{
		\begin{minipage}{\columnwidth}
			\includegraphics[width=\columnwidth]{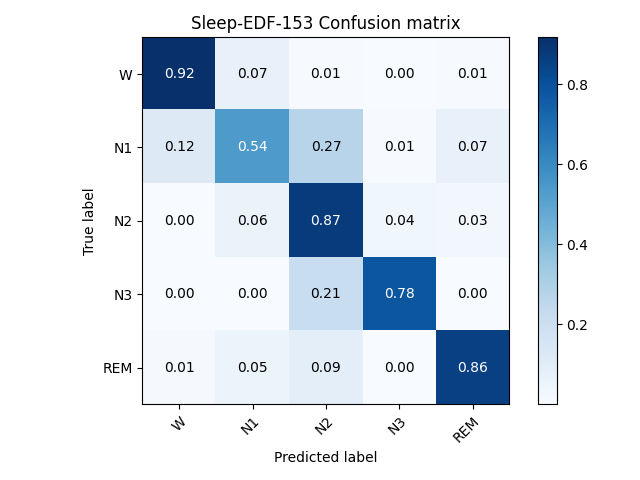} 
		\end{minipage}
	}
	
	\caption{The confusion matrix for Sleep-EDF-39 and Sleep-EDF-153 datasets.}
	\label{fig:cm}
    \end{figure}
	
	\begin{table}[!t]
	\centering
	\caption{The hyperparameters used for all datasets.}
	\resizebox{\columnwidth}{!}{
        \begin{tabular}{ll}
        \toprule
        Parameter              & Value          \\
        \midrule
        Optimizer              & Adam           \\
        Learning rate & 0.001          \\
        $\beta_1$              & 0.9            \\
        $\beta_2$              & 0.999          \\
        $\epsilon$             & $1\times10^{-8}$ \\ \midrule
        Loss function weight & \{1.0, 1.80, 1.0, 1.20, 1.25\} \\
        Regularization       & True                            \\ \midrule
        Input dim & 3000 \\
        Window Size & 20 \\
        Depth & 4 \\
        Up-sampling & bilinear\\ 
        Activations & ReLU \\
        Conv.filters & \{16, 32, 64, 128, 256\} \\ 
        Conv.kernel size & 5\\
        Max-pool kernel size & {10, 8, 6, 4}\\
        Padding & same\\
        Batch Normalization & True \\
        Parameters & $0.9\times 10^6$ \\
        \midrule
        Batch size & 8 \\
        Training epochs & 60 \\
        EarlyStopping patience & 5 \\
        \bottomrule
        \end{tabular}
    }
		\label{tab:hyp}
    \end{table}

\end{document}